\documentclass{article} 
\usepackage[preprint]{colm2026_conference}

\usepackage{microtype}
\usepackage{hyperref}
\usepackage{url}
\usepackage{booktabs}

\usepackage{graphicx}
\usepackage{subcaption}
\usepackage{amssymb}
\usepackage{mathtools}
\usepackage{amsthm}
\usepackage{caption}
\usepackage{multirow}
\usepackage{enumitem}

\usepackage[utf8]{inputenc}
\usepackage[T1]{fontenc}
\usepackage{amsfonts}
\usepackage{nicefrac}
\usepackage[table]{xcolor}

\usepackage{bm}
\usepackage{amsmath}
\usepackage{makecell}
\usepackage{pifont}
\usepackage[capitalize,noabbrev]{cleveref}
\usepackage{xspace}
\usepackage{array}
\usepackage{lineno}

\definecolor{resultgreen}{HTML}{689d6A}
\definecolor{resultred}{HTML}{FB4934}
\definecolor{resultcomparable}{HTML}{a89984}

\definecolor{darkblue}{rgb}{0, 0, 0.5}
\hypersetup{colorlinks=true, citecolor=darkblue, linkcolor=darkblue, urlcolor=darkblue}

\usepackage{amsmath,amsfonts,bm,eqnarray}
\usepackage{cancel}
\usepackage{amsthm}
\usepackage{tikz}
\usetikzlibrary{tikzmark}

\newtheorem{innercustomgeneric}{\customgenericname}
\providecommand{\customgenericname}{}
\newcommand{\newcustomtheorem}[2]{%
  \newenvironment{#1}[1]
  {%
   \renewcommand\customgenericname{#2}%
   \renewcommand\theinnercustomgeneric{##1}%
   \innercustomgeneric
  }
  {\endinnercustomgeneric}
}

\newcustomtheorem{customThm}{Theorem}
\newcustomtheorem{customLemma}{Lemma}
\newcustomtheorem{customCor}{Corollary}
\newcustomtheorem{customProposition}{Proposition}

\def\eqref#1{equation~\ref{#1}}

\def\1{\bm{1}}

\DeclareMathAlphabet{\mathsfit}{\encodingdefault}{\sfdefault}{m}{sl}
\SetMathAlphabet{\mathsfit}{bold}{\encodingdefault}{\sfdefault}{bx}{n}

\renewcommand{\tilde}{\widetilde}

\renewcommand{\frac}{\tfrac}

\renewcommand{\cite}{\citep}

\title{Layer-Parallel Inference Reduces Encrypted Nonlinear Depth in Transformers}

\author{%
  \vspace{-4mm}\\
  \textbf{Ligong Han\textsuperscript{1,2,3}\thanks{Work done while at Red Hat AI Innovation.}~~Kai Xu\textsuperscript{2,3}~~Hao Wang\textsuperscript{2,3}~~Ruijiang Gao\textsuperscript{5}~~Han Gao\textsuperscript{6}~~Akash Srivastava\textsuperscript{3,4}}\\[0.5mm]
  \textsuperscript{1}MBZUAI IFM~~~~~~\textsuperscript{2}Red Hat AI Innovation~~~~~~\textsuperscript{3}MIT-IBM Watson AI Lab\\[0.3mm]
  \textsuperscript{4}Core AI, IBM~~~~~~\textsuperscript{5}University of Texas at Dallas~~~~~~\textsuperscript{6}Iowa State University
}

\begin{document}

\ifcolmsubmission
\linenumbers
\fi

\maketitle

\begin{abstract}
Fully homomorphic encryption (FHE) enables computation on encrypted data, but practical encrypted Transformer inference is bottlenecked by the sequential composition of many nonlinear blocks. We study whether Structured Newton Layer Parallelism (SNLP)~\cite{han2026snlp} can make this inter-layer composition more FHE-friendly: each Transformer block still requires polynomial approximations for operations such as softmax and RMSNorm, but SNLP reduces the layerwise sequential nonlinear depth from $L$ stages to a small number of solver iterations plus linear structured corrections. Using a simulation framework based on Chebyshev polynomial approximations, we measure error accumulation under sequential versus SNLP inference across 8 models and 4 architecture families. On a 0.5B IDN-trained model, SNLP uses $2.65\times$ fewer bootstraps (20 vs.\ 53) and produces encrypted PPL within 1.2\% of sequential's encrypted PPL, while exhibiting lower error amplification ($1.36\times$ vs.\ $1.42\times$). Across all tested models, SNLP has lower amplification than sequential inference. Ablations show that softmax approximation dominates the error budget and CKKS arithmetic noise is negligible in our setting, suggesting that SNLP is complementary to block-level FHE-friendly operator design rather than a replacement for it.
\end{abstract}

\section{Introduction}
\label{sec:intro}

Transformer language models~\cite{vaswani2017attention} are sequential along the layer axis: the hidden state at layer $l+1$ depends on the output of layer $l$. Under fully homomorphic encryption (FHE), this sequential composition is particularly costly. Each Transformer block contains operations that are not native to FHE arithmetic: softmax requires exponentiation, RMSNorm requires inverse square root, and activations require nonlinear evaluation, all of which must be approximated by polynomials when operating on encrypted data~\cite{chen2022thex,hao2022iron,pang2024bolt}. These polynomial approximations introduce per-block errors that compound across $L$ sequential layers, consuming the modulus budget and requiring frequent bootstrapping, the most expensive FHE operation, accounting for 50--86\% of total inference latency~\cite{agrawal2024heap}.

We observe that Structured Newton Layer Parallelism (SNLP)~\cite{han2026snlp} offers a complementary direction for reducing FHE inference cost. SNLP does not make an individual Transformer block FHE-native: softmax, normalization, and activations still require approximation or redesign. Instead, SNLP changes the inter-layer computation graph. Rather than composing $L$ nonlinear block evaluations sequentially, SNLP evaluates suffix layers in parallel over $K$ solver iterations and propagates information across depth using structured Newton-style corrections. For IDN, the correction is purely additive and has zero FHE multiplicative depth; for HCN, it is a small linear mixing over streams. Thus SNLP targets the sequential composition of FHE-unfriendly blocks, not the local block operators themselves.

We introduce the metric \textbf{NFE} (Nonlinear Forward Evaluations) $= (L - N) + K$, where $N$ is the number of parallel suffix layers. In our symbolic CKKS cost model, NFE tracks the FHE bootstrap count closely (ratio $0.99$--$1.02\times$). Using a simulation framework that replaces nonlinear operations with Chebyshev polynomial approximations, we measure how polynomial errors accumulate under sequential versus SNLP inference. Degree-12 Chebyshev softmax approximation degrades sequential PPL by about 42\%; the question is whether SNLP adds to or reduces this shared cost.

Our main findings:
\begin{itemize}[nosep,leftmargin=15pt]
\item SNLP has lower error amplification than sequential across all 8 models at degree 12, and across all tested degrees on the 0.5B IDN model. On that model, SNLP uses $2.65\times$ fewer bootstraps (20 vs.\ 53) and achieves encrypted PPL within 1.2\% of sequential's encrypted PPL, with lower amplification ($1.36\times$ vs.\ $1.42\times$).
\item mHC architectures~\cite{zhu2025hyper,xie2025mhc} are inherently more FHE-friendly ($1.24\times$ amplification vs.\ $1.42\times$ for standard models) because their HC connections are purely linear.
\item FHE-optimal SNLP configurations differ from wallclock-optimal ones: some higher-$K$ configurations trade extra solver depth for lower error while remaining much shallower than the sequential baseline, despite being slower in wall-clock time.
\item Softmax approximation remains the dominant local error source, while CKKS arithmetic noise is negligible at the precisions we test; SNLP should therefore be viewed as complementary to, not a substitute for, FHE-friendly block design.
\end{itemize}

\section{Related Work}
\label{sec:related}

\noindent\textbf{FHE for neural network inference.}
Fully homomorphic encryption enables computation on encrypted data without decryption. The CKKS scheme~\cite{cheon2017ckks} supports approximate floating-point arithmetic, making it the most common choice for neural network inference. CryptoNets~\cite{gilad2016cryptonets} first applied neural networks to encrypted data. Subsequent work on encrypted Transformer inference includes THE-X~\cite{chen2022thex}, Iron~\cite{hao2022iron}, BOLT~\cite{pang2024bolt}, and THOR~\cite{moon2025thor}, which approximate all nonlinear operations (softmax, normalization, activations) with polynomials and report inference times of 30 seconds to 10+ minutes for BERT- or GPT-2-scale models. Beyond inference, HE has also been applied to data valuation and sharing~\cite{yang2025sell}. These approaches focus on improving per-block polynomial approximation quality and reducing per-layer circuit depth. Our work is complementary: rather than changing the per-block approximation, we change the inter-layer computation graph to reduce the number of sequential nonlinear stages.

\noindent\textbf{Parallel solvers and structured recurrences.}
SNLP builds on the view that a sequential computation can be solved as a coupled nonlinear system. DEER~\cite{lim2024parallelizing} applies Newton's method to nonlinear recurrences; later work extends this to MCMC chains~\cite{zoltowski2025parallelizing} and improves scalability with quasi-Newton approximations~\cite{gonzalez2024towards}. Song et al.~\cite{song2021accelerating} frame feedforward computation as parallel equation solving, and Jacobi decoding~\cite{santilli2023accelerating} applies fixed-point iteration to autoregressive translation. Parallel prefix scan~\cite{blelloch1990prefix} underlies efficient recurrent and state-space models~\cite{martin2018parallelizing,gu2022efficiently,gu2024mamba}; SNLP uses the same principle for depthwise correction. Our work rotates these ideas from sequence length to Transformer depth, using structured surrogates that avoid full layer Jacobians.

\noindent\textbf{Depth mixing and efficient inference.}
Residual connections~\cite{he2016deep} are central to deep Transformers; Hyper-Connections and mHC~\cite{zhu2025hyper,xie2025mhc} introduce learned residual-stream mixing, while value residual~\cite{zhou2025value} and $x_0$-style connections~\cite{modded_nanogpt_2024} alter how features persist through depth. Most efficient LLM inference work accelerates per-layer execution through KV caching~\cite{kwon2023efficient}, kernel engineering~\cite{dao2022flashattention}, speculative decoding~\cite{leviathan2023fast,chen2023accelerating}, or early exit~\cite{schuster2022confident}. SNLP targets a different bottleneck: the dependency chain across layers for a fixed token prefix.

\section{Background}
\label{sec:background}

\subsection{FHE for Neural Networks}

FHE schemes such as CKKS~\cite{cheon2017ckks} enable approximate arithmetic on encrypted data. The key cost dimensions are \emph{multiplicative depth} (each ciphertext-ciphertext multiplication consumes one modulus level), \emph{bootstrapping} (refreshing the modulus chain, which is roughly $600\times$ more expensive than a plaintext multiplication~\cite{agrawal2024heap}), and \emph{rotations} (for ciphertext packing and vector operations). Prior FHE-ML work on Transformer inference~\cite{chen2022thex,hao2022iron,pang2024bolt,moon2025thor} reports latencies of 30 seconds to 10+ minutes for a single BERT or GPT-2 forward pass, with polynomial approximation of all nonlinear operations being the primary engineering challenge.

For a Transformer block, the FHE-unfriendly operations are: softmax (requires $\exp$), RMSNorm (requires $\mathrm{rsqrt}$), activation functions (GELU, SiLU, or ReLU), and sigmoid/tanh gates. Linear operations (matrix multiplies, additions, rotations) are natively supported. In a standard $L$-layer Transformer, these nonlinear operations compose sequentially $L$ times, yielding total encrypted nonlinear depth proportional to $L$.

\subsection{SNLP Overview}

Structured Newton Layer Parallelism~\cite{han2026snlp} treats the hidden-state trace $\mathbf{h} = (h_1, \ldots, h_L)$ as the solution of a nonlinear residual equation
\begin{equation}
G_l(\mathbf{h}) = h_l - f_l(h_{l-1}) = 0, \quad l = 1, \ldots, L,
\end{equation}
where $f_l$ is the $l$-th Transformer block. Rather than solving this sequentially, SNLP evaluates the first $S$ layers sequentially (the prefix), then iteratively solves for the remaining $N = L - S$ suffix layers. At iteration $k$, each suffix layer is first evaluated using the current estimate of its input:
\begin{equation}
    \tilde h_l^{(k)} = f_l\!\left(h_{l-1}^{(k)}\right),
    \qquad l=S+1,\ldots,L .
\end{equation}
These evaluations are independent across $l$ and can be batched or fused. SNLP then applies the structured Newton correction:
\begin{equation}
\label{eq:snlp-update}
h_l^{(k+1)} = \tilde{h}_l^{(k)} + A_l^{(k)} \left( h_{l-1}^{(k+1)} - h_{l-1}^{(k)} \right), \quad h_S^{(k+1)} = h_S,
\end{equation}
where $A_l^{(k)}$ is a structured surrogate for the block Jacobian $J_l^{(k)} = \partial f_l / \partial h_{l-1}$. If $A_l^{(k)} = J_l^{(k)}$, this recovers the exact Newton update over depth. SNLP instead chooses $A_l^{(k)}$ so that the correction is much cheaper than evaluating or materializing the true Jacobian, while still propagating information from earlier corrected layer states to later ones.

\textbf{Identity Newton (IDN)} uses $A_l = I$, reducing the correction to additive prefix-sum propagation with zero FHE multiplicative depth. \textbf{HC Newton (HCN)} uses the mHC residual mixing matrix $A_l = H^{\mathrm{res}}_{\mathrm{mlp},l} H^{\mathrm{res}}_{\mathrm{attn},l}$, which is a small $S \times S$ matrix over the stream dimension~\cite{zhu2025hyper,xie2025mhc}. Both surrogates keep the correction cost negligible compared to the nonlinear block evaluations. SNLP-aware training adds an auxiliary loss during pretraining that makes the finite-iteration SNLP solve match the sequential trace, enabling the model to work with the cheap surrogate~\cite{han2026snlp}.

\section{Method}
\label{sec:method}

\subsection{FHE Cost Model and NFE}

We define \textbf{NFE} (Nonlinear Forward Evaluations) as the number of sequential nonlinear stages in the computation graph:
\begin{equation}
\text{NFE}_{\text{seq}} = L, \qquad \text{NFE}_{\text{SNLP}} = (L - N) + K,
\end{equation}
where $N$ is the number of parallel suffix layers and $K$ is the number of SNLP iterations. NFE captures the FHE cost because each nonlinear stage requires polynomial approximation (consuming multiplicative depth and potentially bootstrapping), while the IDN correction adds zero multiplicative depth (it consists only of ciphertext additions and plaintext scalar multiplies).

To verify that NFE is a faithful proxy for actual FHE cost, we build a symbolic CKKS cost model. Each Transformer block contributes a fixed nonlinear multiplicative depth determined by the number of nonlinear operations (two RMSNorm evaluations, one QK-norm, one softmax, one activation, and one sigmoid gate) and the polynomial evaluation method. With 15 usable CKKS levels before bootstrapping, the number of bootstraps scales linearly with NFE. In this symbolic model, NFE tracks the bootstrap count within 1--2\% across all tested configurations (\cref{tab:cost}).

\subsection{HE-Approximation Simulation}

Rather than implementing full CKKS encryption (which would limit us to small models due to the computational overhead), we simulate the effect of FHE-compatible operations by replacing each nonlinear operation with its polynomial approximation in the PyTorch forward pass.

\paragraph{Chebyshev polynomial fitting.}
For each nonlinear function $g(x)$ (e.g., $\exp$, $\sigma$, $\tanh$), we compute a degree-$d$ Chebyshev polynomial approximation on a calibrated interval $[a, b]$. The fitting interval is determined by running 20 batches of validation data through the model and recording the input distribution (min, max, percentiles) for each nonlinear operation at each layer. For softmax, we use the interval $[-20, 0]$ (after max-subtraction for numerical stability), which covers the 99th percentile of attention scores across all layers. At runtime, input values are clamped to the fitting interval before polynomial evaluation. We note that in real CKKS, the max-subtraction used for softmax stability requires comparison circuits, and the normalization step requires reciprocal approximation; these additional costs are not reflected in our NFE metric. Since both sequential and SNLP use identical per-block approximations, this does not affect the relative comparison.

\paragraph{Per-operation approximation.}
We replace each nonlinear operation as follows:
\begin{itemize}[nosep,leftmargin=15pt]
\item \textbf{Softmax}: Chebyshev polynomial approximation of $\exp(x)$ on $[-20, 0]$, followed by renormalization. Degree $d \in \{8, 10, 12, 14\}$ controls the accuracy-depth tradeoff.
\item \textbf{RMSNorm}: Simulated Goldschmidt iteration for $\mathrm{rsqrt}$. Each iteration doubles the precision of the approximation; we model the residual error as controlled multiplicative noise with scale $10^{-(n+1)}$ for $n$ iterations.
\item \textbf{Sigmoid and Tanh}: Chebyshev polynomial approximation on calibrated intervals $[-8, 8]$ and $[-3, 3]$, respectively.
\item \textbf{ReLU$^2$}: Kept exact in our simulation. In real CKKS, $\max(0,x)^2$ is not itself a polynomial and would require either a comparison circuit or a smooth polynomial approximation; we leave this cost out of the simulated approximation error and include only its symbolic depth in the cost model.
\item \textbf{CKKS arithmetic noise}: Optional additive Gaussian noise with standard deviation $\sigma = \text{scale} \cdot 2^{-b}$, where $b$ is the precision bits and scale is the signal magnitude.
\end{itemize}

\paragraph{Implementation.}
We implement two modified forward passes: \texttt{forward\_sequential\_he}, which applies polynomial approximations in all $L$ layers sequentially, and \texttt{forward\_snlp\_he}, which applies polynomial approximations in the $K$ iterations of parallel block evaluation while keeping the IDN correction exact (since it is purely additive). Both paths share the same polynomial approximation code, ensuring that any difference in output is due to the computational structure (sequential vs.\ SNLP), not the approximation quality.

A critical sanity check validates the implementation: with all polynomial degrees set to exact (bypassing approximation), both forward paths produce bit-identical results to their respective standard forward passes (maximum absolute difference $= 0.0$ across all tested models).

\paragraph{Error amplification.}
The key metric is \textbf{error amplification}: $\text{PPL}_{\text{HE}} / \text{PPL}_{\text{exact}}$, where $\text{PPL}_{\text{HE}}$ is the perplexity under polynomial approximation and $\text{PPL}_{\text{exact}}$ is the perplexity with exact operations. This ratio isolates the structural effect of error accumulation: if SNLP has lower amplification than sequential, it accumulates less polynomial approximation error despite computing the same per-block approximations.

\subsection{Models}

We evaluate 8 Nanochat~\cite{nanochat} models spanning 4 architecture families, all with $L = 32$ layers:
\begin{itemize}[nosep,leftmargin=15pt]
\item \textbf{Standard} (3B and 0.5B): baseline and IDN-regularized ($\lambda = \nicefrac{1}{16}$--$0.5$). The 3B model has $n_{\mathrm{embd}} = 2048$ with 16 heads; the 0.5B model has $n_{\mathrm{embd}} = 640$ with 5 heads.
\item \textbf{w/o $x_0$/VE} (0.5B): removes value embeddings~\cite{zhou2025value} and $x_0$ residual connections~\cite{modded_nanogpt_2024}, resulting in fewer nonlinearities per block (no sigmoid gate).
\item \textbf{mHC} (0.5B): manifold-constrained HyperConnections~\cite{xie2025mhc} with 4 streams and HCN correction.
\end{itemize}
All models are trained from scratch on ClimbMix~\cite{diao2025climb}. PPL is evaluated on a fixed validation split at sequence length 2048. For efficiency, most experiments use 200k tokens; we verify that amplification ratios are stable at 1M tokens (within 0.3\%).

\section{Experiments}
\label{sec:experiments}

\paragraph{FHE cost comparison.}
We first evaluate the depth savings predicted by the symbolic CKKS cost model. Because the IDN correction adds no multiplicative depth, the cost is controlled by NFE: replacing 32 sequential nonlinear stages with \texttt{n24-K4} reduces NFE to 12 and bootstraps from 53 to 20, a $2.65\times$ reduction. The \texttt{n24-K8} row illustrates a FHE-specific operating point: it uses more solver iterations than \texttt{n24-K4}, but still has half the bootstrap count of sequential inference while offering better approximation quality in later experiments.

\begin{table*}[t]
\centering
\begin{tabular}{lrrrr}
\toprule
Config & NFE & Bootstraps & Nonlinear Depth & NFE Reduction \\
\midrule
Sequential & 32 & 53 & 768 & $1.00\times$ \\
SNLP \texttt{n8-K4} & 28 & 46 & 672 & $1.14\times$ \\
SNLP \texttt{n12-K2} & 22 & 36 & 528 & $1.45\times$ \\
SNLP \texttt{n20-K4} & 16 & 26 & 384 & $2.00\times$ \\
SNLP \texttt{n24-K4} & 12 & 20 & 288 & $2.67\times$ \\
SNLP \texttt{n24-K8} & 16 & 26 & 384 & $2.00\times$ \\
\bottomrule
\end{tabular}
\caption{Symbolic FHE cost comparison (0.5B model). NFE = nonlinear forward evaluations. Bootstraps estimated with 15 usable CKKS levels per bootstrap. The bootstrap \emph{reduction ratio} is independent of polynomial degree because both sequential and SNLP have the same per-block nonlinear depth; IDN correction adds zero multiplicative depth.}
\label{tab:cost}
\end{table*}

\paragraph{Per-operation ablation.}
The approximation error is highly concentrated in softmax. In \cref{tab:ablation}, approximating RMSNorm or sigmoid alone changes PPL by less than 0.2\%, while the logit softcap tanh contributes 12.8\%. Softmax is the bottleneck: degree-4 and degree-8 approximations are unusable, and even degree 12 raises PPL by 43.7\%. Approximating all operations at degree 12 gives nearly the same degradation as approximating softmax alone, so the remaining experiments focus on how SNLP changes the accumulation of this dominant error source.

\begin{table*}[t]
\centering
\begin{tabular}{lrrr}
\toprule
Approximated Operation & Degree & Sequential PPL & Degradation \\
\midrule
None (exact) & -- & 13.08 & -- \\
RMSNorm only & 1 iter & 13.10 & $+0.2\%$ \\
Sigmoid only & 4 & 13.11 & $+0.2\%$ \\
Tanh only & 4 & 14.76 & $+12.8\%$ \\
Softmax only & 4 & 480.60 & $+36.7\times$ \\
Softmax only & 8 & 168.70 & $+12.9\times$ \\
Softmax only & 12 & 18.80 & $+43.7\%$ \\
All operations & 12 & 18.75 & $+43.4\%$ \\
\bottomrule
\end{tabular}
\caption{Per-operation ablation on the 0.5B IDN model (\texttt{d32s\_idn00625\_npar24\_s3}, 10 batches $\approx$ 20k tokens, seq\_len=2048). Each row approximates one operation while keeping others exact. Absolute PPL differs from \cref{tab:crossmodel} due to smaller eval set; relative degradation is consistent. Softmax dominates the error budget at all polynomial degrees.}
\label{tab:ablation}
\end{table*}

\paragraph{Polynomial degree sweep.}
On the 0.5B IDN model, approximation quality changes sharply between degrees 8 and 10 (\cref{tab:degree}). Degree 8 leaves both sequential and SNLP inference effectively broken, with 12--13$\times$ amplification. From degree 10 onward, SNLP retains a 6--9 pp advantage over sequential, and degrees 12 and 14 are nearly indistinguishable. We therefore use degree 12 as the default in the remaining experiments.

\begin{table*}[t]
\centering
\begin{tabular}{rrrrr}
\toprule
Degree & Seq Amp. & \texttt{n24-K4} Amp. & \texttt{n20-K4} Amp. & SNLP Advantage (\texttt{n24-K4}) \\
\midrule
8 & $12.57\times$ & $11.85\times$ & $12.34\times$ & $0.7$ pp \\
10 & $1.93\times$ & $1.85\times$ & $1.89\times$ & $8.6$ pp \\
12 & $1.42\times$ & $1.36\times$ & $1.38\times$ & $6.3$ pp \\
14 & $1.41\times$ & $1.34\times$ & $1.37\times$ & $6.3$ pp \\
\bottomrule
\end{tabular}
\caption{Error amplification vs.\ polynomial degree (0.5B IDN model, 200k tokens). A phase transition occurs at degree 10. Degrees 12 and 14 give similar results; degree 12 is the practical minimum. SNLP advantage is consistent at all degrees.}
\label{tab:degree}
\end{table*}

\paragraph{CKKS noise robustness.}
Adding simulated CKKS arithmetic noise on top of degree-12 approximation has no visible effect at the precisions we test (\cref{tab:noise}). The amplification values are unchanged, within rounding, from 40-bit precision down to an aggressive 20-bit setting. In this regime, softmax polynomial error is the dominant perturbation; arithmetic noise is roughly four orders of magnitude smaller even at 6-digit precision.

\begin{table*}[t]
\centering
\begin{tabular}{lrrrr}
\toprule
Noise bits & Precision & Seq Amp. & \texttt{n24-K4} Amp. & Seq PPL$_{\text{HE}}$ \\
\midrule
$\infty$ (none) & exact arith. & $1.419\times$ & $1.355\times$ & 19.96 \\
40 & $\sim$12 digits & $1.419\times$ & $1.355\times$ & 19.96 \\
30 & $\sim$9 digits & $1.419\times$ & $1.354\times$ & 19.96 \\
25 & $\sim$7.5 digits & $1.418\times$ & $1.354\times$ & 19.95 \\
20 & $\sim$6 digits & $1.419\times$ & $1.355\times$ & 19.96 \\
\bottomrule
\end{tabular}
\caption{CKKS noise robustness (0.5B IDN model, degree-12). Polynomial approximation error ($\sim$1\% per operation) dominates CKKS arithmetic noise by $\sim$10{,}000$\times$ even at 6-digit precision. All amplification values are identical within rounding.}
\label{tab:noise}
\end{table*}

\paragraph{Cross-model error amplification.}
With the approximation setup fixed at degree 12, we compare sequential and SNLP inference across all 8 models in \cref{tab:crossmodel}. The SNLP configuration selected for each model has lower error amplification than the corresponding sequential run. The largest gap is on the 0.5B IDN-trained model (m4): \texttt{n24-K4} reduces NFE from 32 to 12 and lowers amplification by 6.3 pp. By contrast, baseline models without SNLP-aware training (m3, m5) tolerate only conservative configurations, so their NFE reduction is limited to $1.14\times$ and the amplification improvement stays below 1 pp.

Architecture also matters. The mHC model (m8) has the lowest sequential amplification among 0.5B models ($1.24\times$), close to the 3B model ($1.25\times$). Its HC width and depth connections are linear matrix operations, and the HCN correction is a small $S \times S$ linear transform, so these architectural additions do not introduce extra nonlinear FHE depth. The 3B models are also less sensitive than the standard 0.5B models ($1.25\times$ vs.\ $1.42\times$), consistent with wider hidden states diluting per-element approximation errors.

\begin{table*}[t]
\centering
\resizebox{0.95\linewidth}{!}{%
\begin{tabular}{clllllllll}
\toprule
 & Model & Config & & \multicolumn{2}{c}{PPL (exact)} & \multicolumn{2}{c}{Amp.\ (HE/exact)} & Bootstrap & NFE \\
\cmidrule(lr){5-6}\cmidrule(lr){7-8}
 & & & & Seq & SNLP & Seq & SNLP & & \\
\midrule
\multicolumn{10}{c}{\textbf{Nanochat-3B}} \\
\midrule
m1 & Baseline & \texttt{n8-K4} & & 9.34 & 9.83 & $1.246\times$ & $1.212\times$ & 46 & 28 \\
m2 & IDN Reg. & \texttt{n8-K4} & & 9.34 & 9.37 & $1.270\times$ & $1.267\times$ & 46 & 28 \\
\midrule
\multicolumn{10}{c}{\textbf{Nanochat-0.5B standard}} \\
\midrule
m3 & Baseline & \texttt{n8-K4} & & 13.91 & 14.10 & $1.472\times$ & $1.462\times$ & 46 & 28 \\
m4 & IDN Reg. & \texttt{n24-K4} & & 14.07 & 14.89 & $1.419\times$ & $1.356\times$ & 20 {\small(2.6$\times$)} & 12 {\small(2.7$\times$)} \\
\midrule
\multicolumn{10}{c}{\textbf{Nanochat-0.5B w/o x0/VE}} \\
\midrule
m5 & Baseline & \texttt{n8-K4} & & 15.89 & 15.98 & $1.466\times$ & $1.457\times$ & 46 & 28 \\
m6 & IDN Reg. & \texttt{n24-K4} & & 15.95 & 16.23 & $1.524\times$ & $1.499\times$ & 20 {\small(2.6$\times$)} & 12 {\small(2.7$\times$)} \\
\midrule
\multicolumn{10}{c}{\textbf{Nanochat-0.5B-mHC}} \\
\midrule
m7 & Baseline & \texttt{n8-K4} & & 13.83 & 14.03 & $1.309\times$ & $1.292\times$ & 46 & 28 \\
m8 & HCN Reg. & \texttt{n20-K4} & & 14.19 & 14.92 & $1.238\times$ & $1.217\times$ & 26 {\small(2.0$\times$)} & 16 {\small(2.0$\times$)} \\
\bottomrule
\end{tabular}
}
\caption{Cross-model comparison under degree-12 polynomial approximation (200k tokens, seq\_len=2048). Amplification = PPL$_\text{HE}$ / PPL (each method's own exact PPL as denominator), isolating HE-induced error from solver error. SNLP PPL (exact) is higher than Seq PPL due to finite-$K$ solver error; despite this, SNLP amplification is consistently lower, indicating less polynomial error accumulation. Bootstrap = SNLP bootstrap count from symbolic CKKS cost model. All models have $L = 32$ layers. m4 uses \texttt{batch\_fwd} initialization; all others use \texttt{h0}.}
\label{tab:crossmodel}
\end{table*}

\begin{figure*}[t]
\centering
\includegraphics[width=0.65\textwidth]{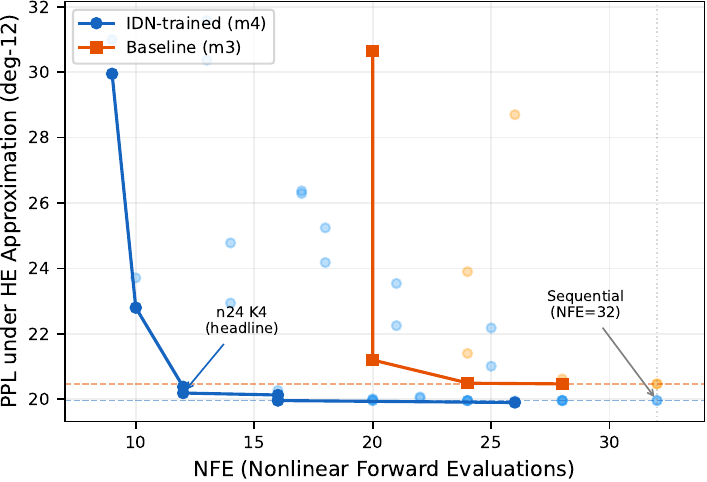}
\caption{NFE vs.\ PPL under degree-12 HE approximation. Each point is one $(N, K, \text{init})$ configuration from a full sweep. The IDN-trained model (blue) dominates the baseline (orange) Pareto frontier, reaching NFE $= 12$ with encrypted PPL within 1.2\% of sequential's. The baseline cannot achieve practical quality below NFE $= 20$, confirming that SNLP-aware training is essential for accessing the low-NFE regime.}
\label{fig:pareto}
\end{figure*}

\paragraph{NFE vs.\ PPL Pareto analysis.}
The full $(N, K)$ sweep in \cref{fig:pareto} separates two effects that are hidden by the single best-configuration table. First, SNLP-aware training shifts the Pareto frontier: the IDN-trained model reaches NFE $= 12$ with encrypted PPL within 1.2\% of sequential's, whereas the baseline cannot achieve practical quality below NFE $= 20$. Second, the useful frontier is narrow. NFE $= 12$--16 gives encrypted PPL within about 1--2\% of sequential's while reducing bootstraps by $2$--$2.65\times$; below NFE $= 10$, quality drops rapidly ($+14$--$50\%$).

The \texttt{n24-K8} point highlights why the FHE frontier differs from the GPU wall-clock frontier. It reaches near-sequential quality ($+0.1\%$) with $2\times$ fewer bootstraps, even though $K = 8$ makes it slower than sequential inference on GPU. For encrypted inference, this is still a plausible operating point because the relevant cost is sequential nonlinear depth rather than ordinary wall-clock time.

\begin{figure*}[t]
\centering
\includegraphics[width=0.65\textwidth]{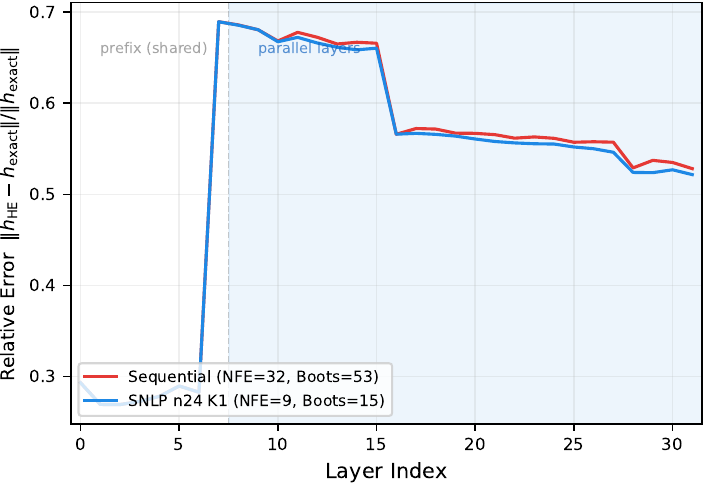}
\caption{Per-layer relative error $\|h_{\text{HE}} - h_{\text{exact}}\| / \|h_{\text{exact}}\|$ under degree-12 polynomial approximation (0.5B IDN model, $N\!=\!24$, $K\!=\!1$). SNLP uses $3.5\times$ fewer bootstraps (15 vs.\ 53) yet achieves comparable per-layer error to sequential inference. Prefix layers (0--7) are shared. In the parallel layers (8--31, shaded), SNLP error stays at or below sequential despite evaluating all suffix blocks in parallel with only one Newton iteration.}
\label{fig:error}
\end{figure*}

\paragraph{Per-layer error accumulation.}
Finally, we inspect where the error difference emerges. For \texttt{n24-K1}, the sequential and SNLP paths share layers 0--7, so their relative errors are identical in the prefix (\cref{fig:error}). The traces diverge in layers 8--31. Sequential inference repeatedly feeds polynomial approximation error into the next nonlinear block, while SNLP evaluates the suffix blocks in parallel within an iteration and propagates corrections through additive linear updates. The final hidden-state error is about 1.2\% lower for SNLP, and the logit-level difference is about 13\%.

\section{Discussion}
\label{sec:discussion}

\paragraph{mHC is most FHE-friendly among 0.5B models.}
The mHC architecture (m8) achieves the lowest sequential amplification ($1.24\times$) among 0.5B models, comparable to the 3B model ($1.25\times$). The HC width and depth connections are purely linear matrix operations that add zero nonlinear FHE depth; the mHC-Newton correction uses a small $S \times S$ matrix multiply (also linear). This suggests that architectures with linear residual-stream mixing are inherently well-suited for encrypted inference, complementing the per-block polynomial approximation improvements pursued by prior FHE-ML work.

\paragraph{FHE-optimal $\neq$ wallclock-optimal.}
The SNLP configuration \texttt{n24-K8} (NFE $= 16$) achieves near-sequential $\text{PPL}_{\text{HE}}$ ($+0.1\%$) with $2\times$ fewer bootstraps, but is \emph{slower} than sequential in wall-clock time because $K = 8$ iterations are compute-heavy. In FHE, cost scales with sequential \emph{depth}, not wall-clock \emph{time}: more iterations at lower depth is cheaper in FHE but slower on GPU. This creates a distinct FHE operating point not captured by the original SNLP speed-quality tradeoff, and suggests that future FHE-oriented SNLP training could specifically optimize for the depth-quality frontier rather than the speed-quality frontier.

\paragraph{Larger models are more robust.}
The 3B models show lower amplification ($1.25\times$) than 0.5B models ($1.42\times$). With wider hidden dimensions ($D = 2048$ vs.\ $640$), per-element polynomial approximation errors are diluted across more dimensions, resulting in less relative perturbation to attention patterns and MLP outputs.

\paragraph{Limitations.}
Our framework simulates polynomial approximation, not actual CKKS encryption. The softmax implementation uses max-subtraction for numerical stability, which requires comparison circuits in real FHE (an additional cost not reflected in our NFE metric). RMSNorm uses controlled noise rather than actual Goldschmidt iteration. These simplifications do not affect the \emph{relative} comparison between sequential and SNLP (both use the same approximations), but mean that absolute FHE cost numbers are estimates. The models are trained at Nanochat scale (0.5B--3B); generalization to larger models remains to be tested, though the trend that larger models have lower amplification is encouraging. Since softmax dominates the approximation error in our ablations, future work should test whether SNLP composes well with architectures that reduce or replace softmax attention, including gated attention~\cite{qiu2026gated}, linear attention~\cite{gu2022efficiently,gu2024mamba,yang2024deltanet}.

\section{Conclusion}
\label{sec:conclusion}

We showed that SNLP provides a complementary direction for FHE-friendly Transformer inference. It does not remove the need to approximate or redesign FHE-unfriendly block operations such as softmax. Instead, it makes the composition of many Transformer blocks more favorable by reducing layerwise sequential nonlinear depth from $L$ stages to $(L - N) + K$ stages, with linear structured corrections. Across 8 models and 4 architecture families, SNLP consistently accumulates less polynomial approximation error than sequential inference. The headline result: SNLP \texttt{n24-K4} uses $2.65\times$ fewer bootstraps (20 vs.\ 53) and achieves encrypted PPL within 1.2\% of sequential's encrypted PPL under degree-12 Chebyshev softmax approximation, while exhibiting lower error amplification ($1.36\times$ vs.\ $1.42\times$).

\bibliography{ref}
\bibliographystyle{colm2026_conference}

\end{document}